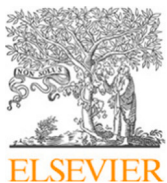
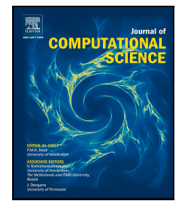

# Physics-informed boundary integral networks (PIBI-Nets): A data-driven approach for solving partial differential equations

Monika Nagy-Huber *, Volker Roth

*Department of Mathematics and Computer Science, University of Basel, Basel, Switzerland*

A B S T R A C T

Partial differential equations (PDEs) are widely used to describe relevant phenomena in dynamical systems. In real-world applications, we commonly need to combine formal PDE models with (potentially noisy) observations. This is especially relevant in settings where we lack information about boundary or initial conditions, or where we need to identify unknown model parameters. In recent years, Physics-Informed Neural Networks (PINNs) have become a popular tool for this kind of problems. In high-dimensional settings, however, PINNs often suffer from computational problems because they usually require dense collocation points over the entire computational domain. To address this problem, we present Physics-Informed Boundary Integral Networks (PIBI-Nets) as a data-driven approach for solving PDEs in one dimension less than the original problem space. PIBI-Nets only require points at the computational domain boundary, while still achieving highly accurate results. Moreover, PIBI-Nets clearly outperform PINNs in several practical settings. Exploiting elementary properties of fundamental solutions of linear differential operators, we present a principled and simple way to handle point sources in inverse problems. We demonstrate the excellent performance of PIBI-Nets for the Laplace and Poisson equations, both on artificial datasets and within a real-world application concerning the reconstruction of groundwater flows.

## 1. Introduction and related work

The study of data-driven partial differential equations (PDEs) is an important area of research owing to its potential to provide insights into complex dynamical systems. In this paper, we focus on linear PDEs with constant coefficients, particularly on the Laplace and Poisson equation, which find many applications in areas such as in hydrogeology [1], electromagnetics [2] or fluid dynamics [3]. Boundary conditions of PDE models are scientifically interesting and ensure the uniqueness of the solution. However, in practice, boundary conditions are often unknown and need to be estimated from noisy measurements and the use of prior knowledge. Additionally, many PDEs used in real-life situations cannot be solved analytically and need numerical solvers instead [4]. Recent research has thus focused on data learning approaches in dynamical systems [5,6]. These methods integrate data into machine learning approaches to improve the accuracy of the data assimilated PDE solution. In conclusion, the study of data-driven PDEs has many practical applications and is an active area of research in mathematics, physics, and engineering.

### 1.1. Data-driven approaches and data assimilation

Data assimilation, as a particular class of inverse problems, shows a growing interest due to the integration of machine learning approaches in recent years [7]. Interpolation techniques and Gaussian processes, as demonstrated in [1,8], are commonly used in data assimilation approaches, but often at the expense of neglecting information of the underlying PDE. In general, classical data assimilation approaches can be categorised mainly into Kalman filters and variational methods [9]. Hybrid deep learning approaches as in [10] combine flexible neural networks with traditional Kalman filters to compute the dynamics. Besides, Physics-Informed Neural Networks (PINNs) [11] have emerged as an attractive alternative to classical methods for solving data-driven PDEs. As a deep learning approach, PINNs incorporate the governing PDE as constraints during training, while leveraging data to improve accuracy and identify parameters. PINNs have been extensively explored for solving various forward and inverse problems as for example in [12]. However, recent findings [13,14] indicate that Physics-Informed Neural Networks (PINNs) have limited extrapolation capabilities due to the large number of collocation points required, especially in high dimensional spaces. Therefore, there is a need to either find better sampling strategies as proposed in [15] or lowering the dimensionality of the problem by incorporating reduced-order modelling into PINNs as shown in [16]. Each of these variations involves tailoring PINNs to suit specific applications. Nonetheless, the foundation lies in the universally applicable PINNs, which makes them also to our preferred method for comparisons.

Since the Laplace and Poisson equations are one of the most important PDEs in many fields of engineering and science, there is still ongoing research, especially in combination with neural network approaches [17]. On the one hand, Convolutional Neural Networks (CNNs) are used to solve the Poisson equation [18] on a Cartesian mesh. Follow up research from [19,20] conducted a study on a fast solver for Poisson's equation based on CNN. On the other hand, PINNs are powerful solvers especially in inverse problems [21] and still a state of the art approach. Similar approaches to the CNNs are Maria Antony et al. [22] and Abdulkadirov et al. [23] that solve the Laplace and Poisson equation on a mesh, and integrating automatic differentiation for speed up [24].

### 1.2. Approaches with boundary integral equations

In contrast to PINN approaches that require dense collocation points over the entire domain, boundary integral equation methods only require dense integration points at the boundary. This reduces the problem dimension by one. Both, classical methods such as the boundary element method (BEM) described in [25] and deep learning approaches [26,27] require boundary conditions as a prior. However, in many real-world applications, boundary conditions are commonly unknown and need therefore be estimated as in [28]. To unite both advantages of leveraging data measurements with PDE dynamics as in PINNs, and reducing the problem dimension by one as in BEMs, we introduce Physics-Informed Boundary Integral Networks (PIBI-Net) as a data-driven approach based on integral equations to target this challenge.

This paper is organised as follows: In Section 2 we provide a brief introduction to boundary integral equations, which constitute the theoretical foundation of the PIBI-Net. In particular, we present the boundary integral equations for the Laplace as well as the Poisson equation with unknown point sources, and derive an analytical formula to handle these point sources in an inverse problem setting. We then showcase PIBI-Net for the Laplace and Poisson equation on synthetic data as well as on a real-world dataset in Section 3. Lastly, we compare PIBI-Net to PINNs in Section 4 and summarise our findings in Section 5.

## 2. Methods

PIBI-Net is based on the boundary integral representation of linear PDEs with constant coefficients

$$\mathcal{L}u = f. \tag{1}$$

In this equation, $\mathcal{L}$ represents the PDE operator, $u$ the solution, and $f$ the source term. In our setting, the objective is to find the optimal solution $u : \Omega \subseteq \mathbb{R}^d \to \mathbb{R}$ that assimilates the Poisson equation

$$\begin{cases} \Delta u(\mathbf{x}) = f(\mathbf{x}) & \text{in } \Omega \\ u(\mathbf{x}_i) = u_i, & i = 1, \dots, N, \quad N \in \mathbb{N} \end{cases} \tag{2}$$

within the domain $\Omega$ to a set of data measurements $\mathbf{u}^{\text{data}} := \{u_i\}_{i=1}^N$ in $\mathbb{R}^{N \times 1}$ at $\mathbf{X}^{\text{data}} := \{\mathbf{x}_i\}_{i=1}^N$ in $\mathbb{R}^{N \times d}$ with unknown point sources, whereas the PDE operator is $\mathcal{L} = \Delta := \sum_{i=1}^n \frac{\partial^2}{\partial x_i^2}$. Our proposed PIBI-Net method enables learning both the solution $u$ of Eq. (2) and the unknown source term $f$.

### 2.1. Boundary integral equation for the Laplace and Poisson equation

In order to lay a foundation for the architecture of PIBI-Net, we review key concepts from potential theory. According to the Malgrange–Ehrenpreis theorem [29,30], every linear PDE (1) with constant coefficients has a fundamental solution, which is commonly known as Green's function. The fundamental solution $G$ for a linear partial differential operator $\mathcal{L}$ is defined as

$$\mathcal{L}G = \delta, \tag{3}$$

where $\delta$ denotes the Dirac delta function. In particular, for the Laplace or Poisson equation (2) in dimension $d$, the fundamental solution is given by

$$G(\mathbf{x}, \mathbf{y}) := \begin{cases} -\frac{1}{2\pi} \log(\|\mathbf{x} - \mathbf{y}\|_2) & \text{for } d = 2 \\ \frac{1}{4\pi \|\mathbf{x}-\mathbf{y}\|_2} & \text{for } d \geq 3, \end{cases} \tag{4}$$

where $\mathbf{x}, \mathbf{y} \in \Omega$ and $\|\cdot\|$ denotes the Euclidean norm. Fundamental solutions for various PDEs, including Laplace, Poisson, heat, wave, Stokes, and Helmholtz equations are available in [25,31] as well as in [32]. The gradient of the fundamental solution (4) with respect to $\mathbf{y}$ on the boundary $\partial \Omega$ can be analytically derived with the chain rule as

$$\nabla_{\mathbf{y}} G(\mathbf{x}, \mathbf{y}) = \begin{cases} \frac{(\mathbf{x}-\mathbf{y})}{2\pi \|\mathbf{x}-\mathbf{y}\|_2^2} & \text{for } d = 2, \\ \frac{(\mathbf{y}-\mathbf{x})}{4\pi \|\mathbf{x}-\mathbf{y}\|_2^3} & \text{for } d \geq 3. \end{cases} \tag{5}$$

Let $\int_{\partial \Omega} (*) \, ds_{\mathbf{y}}$ indicates the surface integral over the boundary of $\Omega$. For non-homogeneous PDEs like the Poisson equation, considering an external force $f$, we define the source potential as follows:

$$\mathcal{F}[G, f](\mathbf{x}) := \int_{\Omega} G(\mathbf{x}, \mathbf{y}) f(\mathbf{y}) \, d\mathbf{y}, \quad \text{for } \mathbf{x} \in \Omega. \tag{6}$$

Here, $\mathbf{z}$ and $\mathbf{y}$ denote points on the boundary $\partial \Omega$, and $\mathbf{n}_{\mathbf{y}}$ represents the outward-directed normal vector at $\mathbf{y}$. In the case of the homogeneous Laplace equation, where the right-hand side in (2) is zero, the source potential vanishes.

Let us denote the homogeneous part of the solution at the boundary by $h$. For the Laplace and Poisson equations, we recall from [25] the single layer potential defined as

$$\mathcal{S}[G, h](\mathbf{x}) := \int_{\partial \Omega} G(\mathbf{x}, \mathbf{y}) \frac{\partial h}{\partial \mathbf{n}_{\mathbf{y}}}(\mathbf{y}) \, ds_{\mathbf{y}} \quad \text{for } \mathbf{x} \in \Omega, \tag{7}$$

which retains its property on the boundary $\partial \Omega$

$$\lim_{\mathbf{x} \to \mathbf{z}} \mathcal{S}[G, h](\mathbf{x}) = \mathcal{S}[G, h](\mathbf{z}). \tag{8}$$

Further, the double layer potential is defined as

$$\mathcal{D}[G, h](\mathbf{x}) := \int_{\partial \Omega} \frac{\partial G}{\partial \mathbf{n}_{\mathbf{y}}}(\mathbf{x}, \mathbf{y}) h(\mathbf{y}) \, ds_{\mathbf{y}} \quad \text{for } \mathbf{x} \in \Omega, \tag{9}$$

with the property on the boundary $\partial \Omega$ given by

$$\lim_{\mathbf{x} \to \mathbf{z}} \mathcal{D}[G, h](\mathbf{x}) = \int_{\partial \Omega} \frac{\partial G}{\partial \mathbf{n}_{\mathbf{y}}}(\mathbf{z}, \mathbf{y}) h(\mathbf{y}) \, ds_{\mathbf{y}} - \frac{1}{2} h(\mathbf{z}). \tag{10}$$

The outer normal derivatives can be easily calculated by the dot product $\frac{\partial (*)}{\partial \mathbf{n}_{\mathbf{y}}} = \nabla_{\mathbf{y}}(*) \cdot \mathbf{n}_{\mathbf{y}}$, where $(*)$ stands for $h$ or $G$. For sake of simplicity, we write $(*)_{\mathbf{n}_{\mathbf{y}}} := \frac{\partial (*)}{\partial \mathbf{n}_{\mathbf{y}}}$. Recalling [25], the boundary integral representation formula based on (4)–(10) for any solution of the Laplace or Poisson equation (2) is given by

$$u(\mathbf{x}) = \begin{cases} \left(\mathcal{S}[G, h] - \mathcal{D}[G, h]\right)(\mathbf{x}) + \mathcal{F}[G, f](\mathbf{x}) & \text{for } \mathbf{x} \in \Omega \\ \lim_{\mathbf{x}' \to \mathbf{x}} \left(\mathcal{S}[G, h] - \mathcal{D}[G, h]\right)(\mathbf{x}') + \mathcal{F}[G, f](\mathbf{x}) & \text{for } \mathbf{x} \in \partial \Omega, \end{cases} \tag{11}$$

where $\mathbf{x}'$ denotes a point in $\Omega$.

*2.2. Physics-Informed Boundary Integral Network (PIBI-Net)*

PIBI-Net is a deep learning architecture that aligns observations to the linear PDE with constant coefficients. Its construction is based on the boundary integral representation (11). In the non-homogeneous Poisson equation, a known right hand-side can be generally calculated analytically or simplified by a spline approximation as shown in [33]. However, in this paper, we showcase it for unknown point sources in an inverse problem setting. In general, a point source can be represented by a Dirac delta function with magnitude $c_0 \in \mathbb{R}$. Consequently, the integral (6) over the entire domain $\Omega$ simplifies to the point source potential

$$\mathcal{F}[G, f](\mathbf{x}) = \sum_{i=1}^{M} c_{0_i} G(\mathbf{x}, \mathbf{y}_{0_i}), \quad (12)$$

with $M$ point sources at locations $\mathbf{y}_{0_1}, \ldots, \mathbf{y}_{0_M}$ within the computational domain $\Omega$ and magnitudes $c_{0_1}, \ldots, c_{0_M} \in \mathbb{R}$.

To implement PIBI-Net, we employ a multi-layer perceptron (MLP) to learn an approximation $\hat{h} : \mathbf{x} \to \hat{h}(\mathbf{x}|\mathbf{w})$ with weights $\mathbf{w}$ of the homogeneous solution and, if present but unknown, the parameters of $\hat{f}$ from the approximation of the right-hand side term given by Eq. (12). Let $X^{int} := \{\mathbf{x}_j^{int}\}_{j=1}^{I}$ in $\mathbb{R}^{I \times d}$ with $I \in \mathbb{N}$ be a set of integration points on the boundary $\partial \Omega$. Furthermore, denote $V := \int_{\partial \Omega} ds_\mathbf{y}$ as the volume of the computational domain boundary $\partial \Omega$. The boundary integral representation in Eq. (11) can then be approximated by Monte Carlo integration as

$$\hat{u}(\mathbf{x}) = \begin{cases} \frac{V}{I} \sum_{j=1}^{I} \left( G\left(\mathbf{x}, \mathbf{x}_j^{int}\right) \hat{h}_{\mathbf{n}_{\mathbf{x}_j^{int}}} - G_{\mathbf{n}_{\mathbf{x}_j^{int}}}(\mathbf{x}, \mathbf{x}_j^{int}) \hat{h}(\mathbf{x}_j^{int}) \right) \\ \qquad + \sum_{i=1}^{M} \hat{c}_{0_i} G(\mathbf{x}, \hat{\mathbf{y}}_{0_i}), & \mathbf{x} \in \Omega \\ \frac{V}{I} \sum_{j=1}^{I} \left( G\left(\mathbf{x}, \mathbf{x}_j^{int}\right) \hat{h}_{\mathbf{n}_{\mathbf{x}_j^{int}}} - G_{\mathbf{n}_{\mathbf{x}_j^{int}}}(\mathbf{x}, \mathbf{x}_j^{int}) \hat{h}(\mathbf{x}_j^{int}) \right) \\ \qquad + \frac{1}{2} \hat{h}(\mathbf{x}) + \sum_{i=1}^{M} \hat{c}_{0_i} G(\mathbf{x}, \hat{\mathbf{y}}_{0_i}) & \mathbf{x} \in \partial \Omega, \end{cases} \quad (13)$$

where the integration points $\{\mathbf{x}_j^{int}\}_{j=1}^{I}$ should be chosen uniformly over the boundary. Let $\hat{u}(X^{data}) := \{\hat{u}(\mathbf{x}_i)\}_{i=1}^{N}$ in $\mathbb{R}^{N \times 1}$ denote the boundary integral approximation over the set of measurements. In the presented PIBI-Net approach, the MLP solely learns the homogeneous part of the solution $\hat{h}$, that is subsequently utilised to approximate the boundary integral equations using the formulation given in Eq. (13). In general, the MLP can be in practice any feedforward type of network architecture, such as a fully connected neural network. Automatic differentiation in common deep learning frameworks like PyTorch [34] and TensorFlow [35] enables the MLP to learn both the homogeneous part of the solution $\hat{h}$ and its derivative.

To train PIBI-Net, the loss function $\mathcal{L}$ consists in general of an observation loss and a boundary loss. The observation loss $\mathcal{L}^{obs.} := \left\| \hat{u}(X^{data}) - \mathbf{u}^{data} \right\|_2^2$ ensures that PIBI-Net model aligns with the observations, whether they are within the computational domain $\Omega$ or on its boundary $\partial \Omega$. Unlike the PINN approach, where collocation points are always necessary, PIBI-Net may require them only if the boundary function is partially known or if there exists measurements at the boundary. In such cases, collocation points $X^{coll} := \{\mathbf{x}_j^{coll}\}_{j=1}^{K}$ are chosen on the boundary $\partial \Omega$ and the boundary loss calculated as $\mathcal{L}^{boundary} := \left\| \hat{u}(X^{coll}) - \hat{h}(X^{coll}) \right\|_2^2$ encodes the function value at the boundary. Here, $\hat{u}(X^{coll}) := \{\hat{u}(\mathbf{x}_j)\}_{j=1}^{K}$ in $\mathbb{R}^{K \times 1}$ denotes the boundary integral approximation over the set of collocation points given by Eq. (13). The loss function for PIBI-Net can be then formulated as

$$\mathcal{L}oss := \mathcal{L}^{obs.} + \lambda \mathcal{L}^{physics} = \left\| \hat{u}(X^{data}) - \mathbf{u}^{data} \right\|_2^2 + \lambda \left\| \hat{u}(X^{coll}) - \hat{h}(X^{coll}) \right\|_2^2 \quad (14)$$

where $\lambda \in \mathbb{R}$ denotes a parameter weighting the importance of the boundary loss. However, in many real-world settings the boundary $\partial \Omega$ can be chosen arbitrarily, such that all observations lie inside the computational domain $\Omega$. This simplifies the loss term to $\mathcal{L}oss = \mathcal{L}^{obs.}_{inside}$.

Minimisation of the loss function can be achieved via a gradient-based optimiser like Adam [36]. A visualisation of the architecture for collocation points and observations can be seen in Fig. 1.

We want to point out that the source term $\hat{f}$ needs to be learnt as an additional parameter only if it is not precisely known. In many real-world applications, we typically have prior knowledge of the number $M$ of point sources and a general understanding of their approximate locations. However, in an inverse problem setting where the point sources are unknown, we only need to learn the magnitudes $c_{0_1}, \ldots, c_{0_M}$ and, if necessary, their exact locations $\mathbf{y}_{0_1}, \ldots, \mathbf{y}_{0_M}$, as stated in (12). After the training procedure of PIBI-Net, we still need to compute the boundary integral representation defined by (13) at each point $\mathbf{x}$ for which we wish to evaluate the PIBI-Net solution. Despite this additional computation, the use of Monte Carlo integration nevertheless works efficiently in practice.

## 3. Experiments and results

We showcase the power of the PIBI-Net in comparison to PINNs. In Section 3.1 we demonstrate PIBI-Net on a synthetic dataset for the Laplace equation. In Section 3.2 we extend the synthetic dataset by adding randomly generated point sources to the Poisson equation. To cover the relevance of solving data-driven non-homogeneous PDEs with unknown point sources, we apply PIBI-Net to real-world measurements for groundwater flows in Section 3.3.

*3.1. Laplace equation based on synthetic data*

In this experiment, we compare PIBI-Net with PINNs for the two-dimensional Laplace equation and use the finite difference method as a ground truth reference on an equidistant mesh with mesh size $\Delta \mathbf{x} = 0.2$. We constructed the dataset by $N \in \mathbb{N}$ randomly sampled measurements from the solution obtained by the finite difference method on the Laplace equation $\Delta u(\mathbf{x}) = 0$ in $\Omega := (-1, 1)^2$, with the Dirichlet boundary conditions given as

$$\left. \begin{aligned} u(x_1 = -1, x_2) = u(x_1 = 1, x_2) &= \sin\left(\frac{5}{2} \pi x_2\right) \\ u(x_1, x_2 = -1) &= -1 \\ u(x_1, x_2 = 1) &= 1. \end{aligned} \right\} \quad (15)$$

The variable $\mathbf{x} = (x_1, x_2)$ denotes a point in the closure $\bar{\Omega} := [-1, 1] \times [-1, 1]$ as a subset of $\mathbb{R}^2$. As in many real-world applications, we only have given data measurements and therefore treat the boundary conditions as unknowns for our data assimilation problem setting. To showcase the behaviour of outliers, we have additionally placed a data point with magnitude 2 at the center. Furthermore, we added Gaussian noise $\mathcal{N}(\mu = 0, \sigma^2 = 0.2^2)$ to the dataset to mimic real-world scenarios.

We obtained the results in Fig. 2 by training PIBI-Net for the two-dimensional Laplace equation based on $N = 50$ data measurements with 200 integration points uniformly sampled over a chosen boundary of $[-1 - \varepsilon, 1 + \varepsilon]^2$ with $\varepsilon = 0.1$. We trained the PINN accordingly with 200 collocation points over the entire domain $[-1, 1]^2$ such that the functionality of the PINN method is not restricted. For the PINN, we defined the loss function as $\|\hat{u}_{PINN}(X_{data}) - U_{data}\|_2^2 + \lambda_{physics} \|\hat{u}_{PINN}(X_{coll})\|_2^2$, where $\hat{u}_{PINN}$ represents the Laplace solution obtained by the network and $\lambda_{physics}$ represents a weighting factor to balance both loss terms. The mean and standard deviation of the mean absolute error calculations evaluated on a regular grid with grid size 0.02 is presented in Table 1.

In many real-world data assimilation applications, the primary interest lies not in the boundary function but in the solution of the Laplace or Poisson equation itself. Therefore, the computational boundary can often be chosen arbitrarily. We have found that it is the best to choose a boundary that is slightly larger than the desired computational domain to avoid fluctuations at the boundary itself caused by the jump term in the boundary integral formulation (11). In this way, all data measurements lie within our chosen boundary and we only need to consider the data loss term for points inside the domain given (14). In both approaches we used a fully connected neural network with three hidden layers containing 64 neurons each with hyperbolic tangent activation functions, such that both methods has the same number of network parameters. We trained both networks with Adam [36] as a gradient-based optimiser for $10'000$ iterations. For the loss function, we used the mean squared error metric in both approaches.

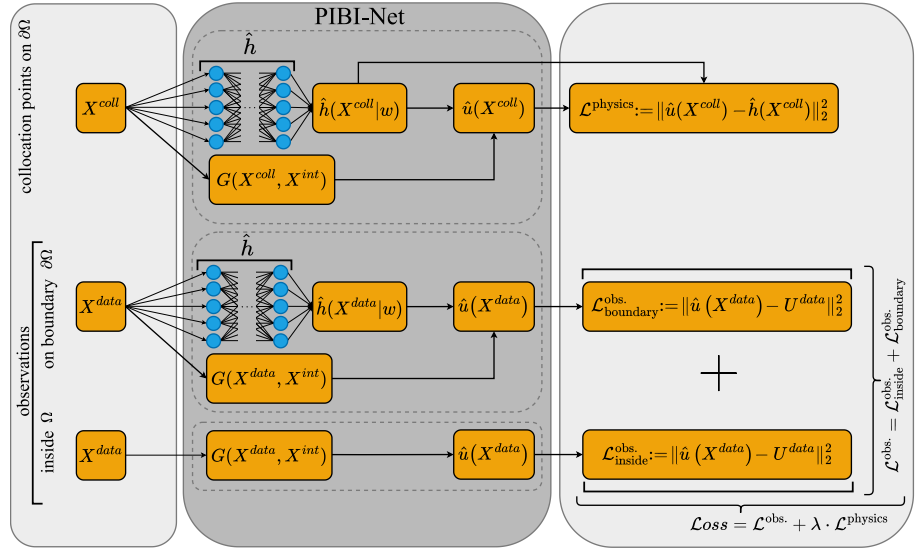

**Fig. 1.** Visualisation of the PIBI-Net architecture. The approximated solution $\hat{u}$ is calculated by Eq. (13), while the MLP optimises homogeneous part of the solution $\hat{h}$.

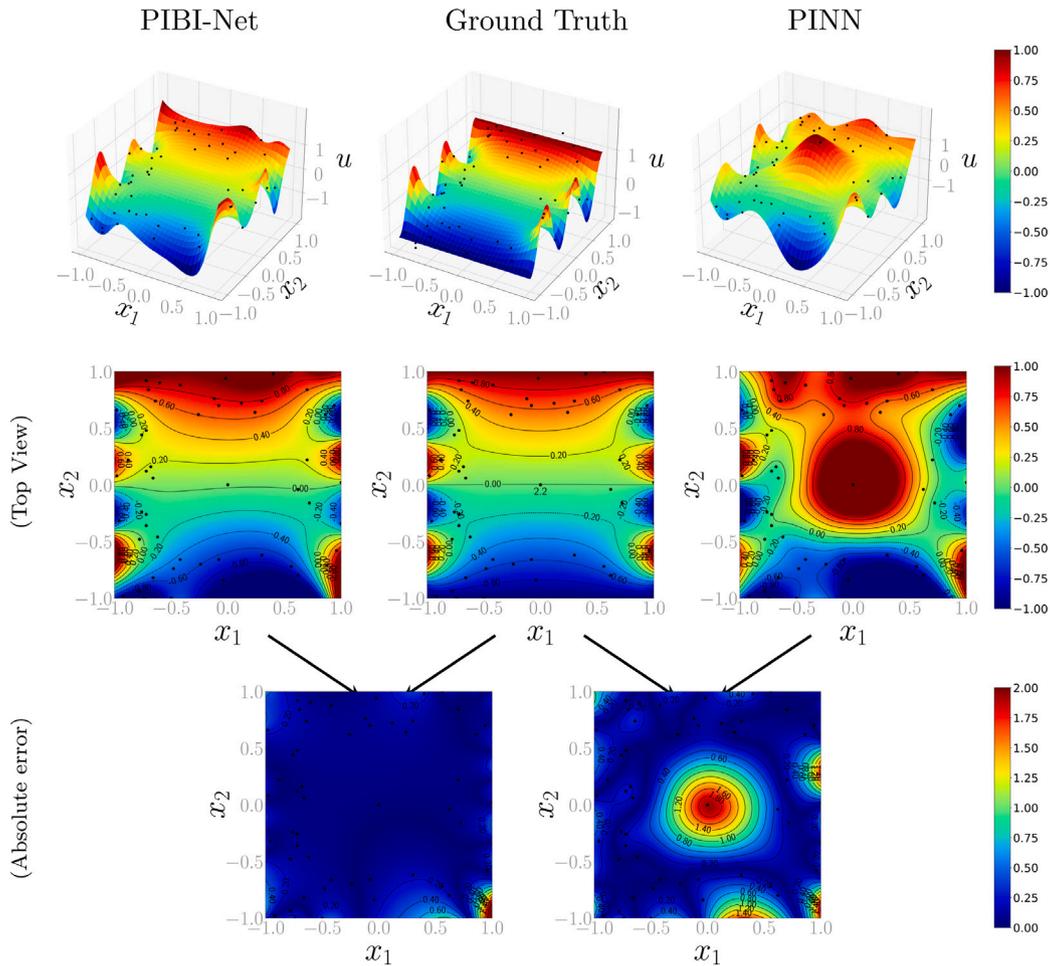

**Fig. 2.** Visualisation of the 2D toy example for solving the Laplace equation using $N = 50$ measurements randomly sampled over the domain $\Theta := (-1, 1)^2 \setminus [-0.6, 0.6]^2$ from the ground truth solution based on (15) and an outlier placed to the center. The data measurements are indicated as black dots, with PIBI-Net listed in the first column and PINNs in the last column. Through cross-validation, we set $\lambda_{\text{physics}} = 0.0001$ for the PINN. We display the three-dimensional representations (top row), the projection to the $x_1x_2$-plane (middle row) and the absolute errors with respect to the ground truth (bottom row). We added contour lines to the two dimensional plots to highlight the differences. The mean absolute errors are 0.109 for PIBI-Net and 0.369 for the PINN.

**Table 1**
Comparison of PIBI-Net with PINN to the ground truth calculated by a finite difference solver. The error is presented as the mean and standard deviation of the mean absolute error, computed over ten randomised datasets using a regular evaluation grid.

|  | Dataset on $\Theta$ | | Dataset on $\Omega$ | |
| --- | --- | --- | --- | --- |
| Sample size | N = 50 | N = 200 | N = 100 | N = 500 |
| PIBI-Net | 0.156 ± 0.036 | 0.051 ± 0.009 | 0.109 ± 0.033 | 0.041 ± 0.009 |
| PINN | 0.258 ± 0.135 | 0.286 ± 0.096 | 0.126 ± 0.025 | 0.062 ± 0.013 |

To analyse and compare the accuracy of the two methods across different datasets, we performed a point-wise subtraction of the ground truth from both the PIBI-Net and the PINN solution based on a regular evaluation grid with grid size 0.02. Table 1 presents a comparison between PIBI-Net and PINN for different sample sizes $N$, with 200 integration or collocation points used for all calculations. For sample sizes $N = 50$ and $N = 200$, measurements were randomly sampled from the domain $\Theta := (-1,1)^2 \setminus [-0.6, 0.6]^2$, while for $N = 100$ and $N = 500$ measurements were sampled from the entire domain $\Omega = (-1,1)^2$. Additionally, in all experiments for the Laplace equation we added an outlier datapoint with a magnitude of 2 to the center. In case of the PINN approach, the parameter $\lambda_{\text{physics}}$ was determined through cross-validation across a range of values, specifically $\lambda_{\text{physics}} \in \{10^k \mid k = -8, -7, -6, -5, -4, -3, -2, -1, 0, 1\}$.

### 3.2. Poisson equation with point sources based on synthetic data

We compare PIBI-Net with PINNs for the two-dimensional Poisson equation with randomly generated point sources as an unknown right hand-side, and use the finite difference method as a ground truth reference on an equidistant mesh with mesh size $\Delta x = 0.2$. More specifically, we want to solve the Poisson equation (2) with $f(\mathbf{x}) = \sum_{i=1}^M c_{0_i} \delta(\mathbf{x} - \mathbf{y}_{0_i})$. The Poisson equation with point sources represented by Dirac delta functions poses a challenge for the PINN method due to the resulting singularities. To address this issue in our experiment, we approximate the Dirac delta function by a Gaussian distribution

$$d_i(\mathbf{x}) = \frac{1}{\sigma\sqrt{2\pi}} e^{-\frac{1}{2}\left(\frac{\mathbf{x}-\mathbf{y}_{0_i}}{\sigma}\right)^2} \quad (16)$$

with standard deviation $\sigma := 0.001$. Eq. (16) enables us to employ a formulation for the PINN similar to the one used for our PIBI-Net by the superposition of the point sources

$$f(\mathbf{x}) = \sum_{i=1}^M c_{0_i} d_i(\mathbf{x}). \quad (17)$$

For the ground truth solution using finite differences, we randomly selected five point source locations within the domain $\Omega$, with magnitudes drawn from a uniform distribution ranging from $-5$ to $5$. From this configuration, we generated a dataset of 50 samples distributed throughout the domain $\Omega$. To simulate real-world conditions, we added Gaussian noise $\mathcal{N}(\mu = 0, \sigma^2 = 0.2^2)$ to the dataset. Following the same network settings as in Section 3.1, we enhanced both network architectures by incorporating additional learnable parameters for the location and magnitude of the unknown sources. In both approaches we used 200 integration or collocation points.

In Fig. 3 we illustrate the comparison of PIBI-Net and PINN approaches in an inverse problem setting where the point source locations and the magnitudes are added as additional learnable parameters to the network. While PIBI-Net is able to learn the locations and magnitudes of the point sources given the sample size of $N = 80$, we faced challenges in the PINN approach that we discuss in Section 4.

### 3.3. Groundwater level measurements in Memphis aquifer in steady-state

Steady-state flow of groundwater in a homogeneous aquifer can be described through the Laplace equation. However, in real-world scenarios, it is challenging to identify areas where this equation is fulfilled owing to the widespread presence of pump wells. In order to handle pump wells correctly, we need to include additional point sources, which leads to the Poisson equation (2). Groundwater levels therefore becomes important for studying historical water-level changes.

In this real-world experiment, we simulated the altitudes of the potentiometric surface of water for the Memphis (Tennessee) area in the US for the fall 2000 based on 64 well measurements. The dataset is based on the water-level altitude measurements presented in [1]. We used the same training settings as described in Section 3.2 with 500 integration points. Fig. 4 visualises the flow lines of the gradient field of the solution, created using the streamplot function from the matplotlib library. This approach is equivalent to employing the conjugate gradient method to calculate the streamlines.

### 4. Discussion

In Section 3, we compared PIBI-Net to PINNs for solving the Laplace and Poisson equations in a data assimilation setting. We analysed the performance of PIBI-Net by comparing the mean absolute error for different number of sample sizes in Table 1. The results clearly indicate that PIBI-Net significantly outperforms PINNs, particularly in scenarios involving outliers and clustered noisy measurements. Furthermore, we demonstrated in Figs. 3 and 4 that PIBI-Net can easily deal with unknown point sources in inverse problem settings. However, we found that while PIBI-Net offers some advantages, it has its limitations as well.

- Our results in Table 1 demonstrate that our proposed PIBI-Net can better reconstruct the ground truth solution than PINNs when applied to data assimilation settings. The PIBI-Net solution consistently exhibits a smaller mean absolute error compared to the ground truth across all settings, including those with measurements near the boundary and those distributed across the entire computational domain. This under-performance of the PINN relies on the $\lambda_{\text{physics}}$ hyperparameter as well as on the need to approximate the Dirac delta function for point sources.
- The balancing parameter $\lambda_{\text{physics}}$ remains pivotal for fitting data and satisfying the Laplace or Poisson equation. This can be challenging and time-consuming. While settings with too large $\lambda_{\text{physics}}$ insufficiently assimilates data, in other settings with too small $\lambda_{\text{physics}}$ as shown in Fig. 2, PINNs do not fulfil the PDE. In contrast, PIBI-Net requires this hyperparameter only if the boundary function is of particular interest. In most cases, the boundary can be chosen arbitrarily, making the loss function independent of such weighting. Moreover, PIBI-Net always satisfies the PDE by construction, regardless of this hyperparameter.
- As shown in our experiments in Fig. 3, PINNs encounter challenges with unknown point source locations and magnitudes. In settings with a small number of collocation points, it is unlikely that a sufficient amount of collocation points would deviate from zero in the right-hand side term given by Eq. (17). This results in making PINNs unaware of the point source's presence. Consequently, the loss term becomes inconsistent, as it either focuses on fitting the data or attempts to satisfy the Laplace equation.

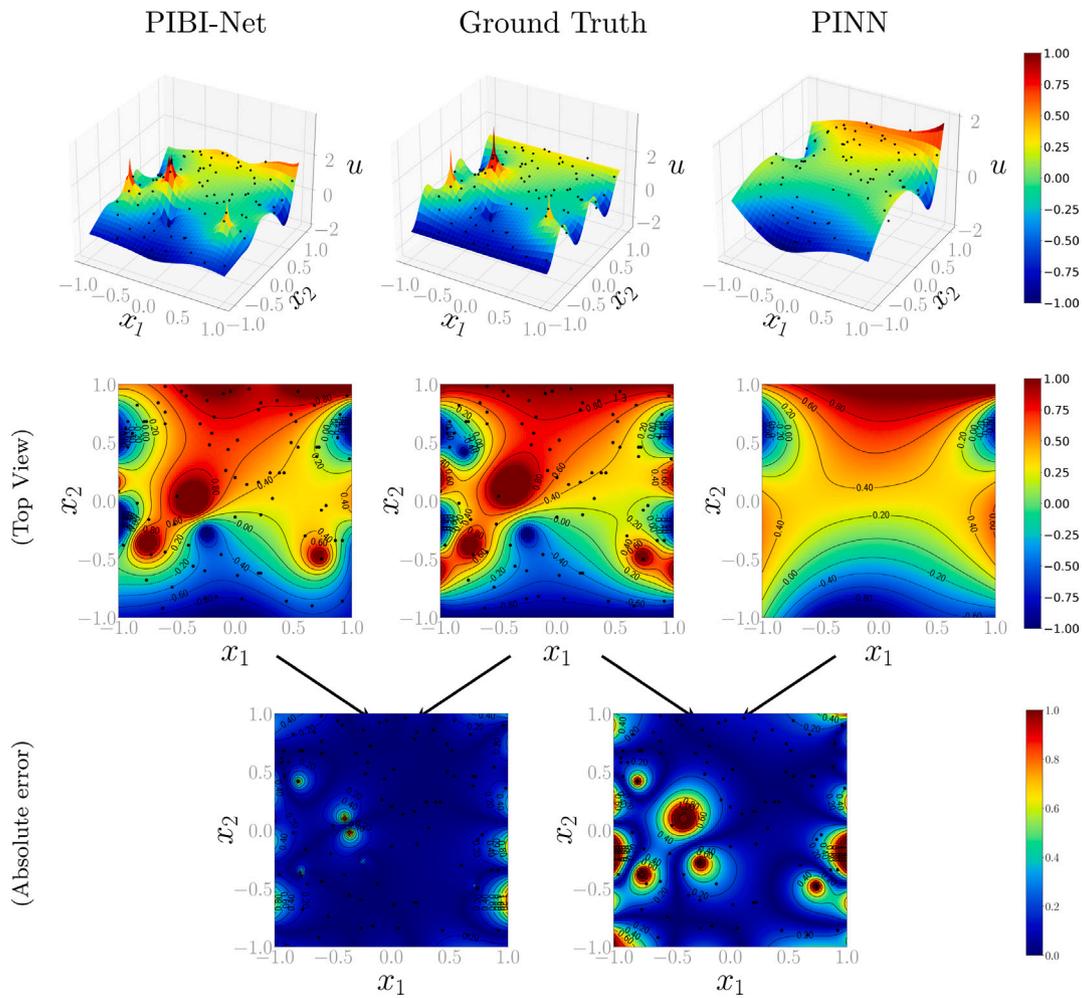

**Fig. 3.** Visualisation of the 2D toy example for solving the Poisson equation using $N = 80$ measurements randomly sampled over the whole domain $\Omega := (-1, 1)^2$. Black dots represent data measurements. The PIBI-Net solution is listed in the first column, the ground truth calculated by finite differences in the middle column and the PINN solution is listed in the last column. The figure shows three-dimensional representations (top row), projections onto the $x_1 x_2$-plane (middle row), and the absolute errors compared to the ground truth (bottom row) and a regular evaluation grid with grid size 0.02. Contour lines in the two dimensional plots highlight the differences. Through cross-validation, we set $\lambda_{\text{physics}} = 0.01$ for the PINN. The mean absolute errors are 0.131 for PIBI-Net and 0.199 for the PINN.

- With known source terms, the particular solution $v$ of the Poisson equation (2) can be calculated. The Poisson equation can then be rewritten to a Laplace equation $\Delta \hat{u} = 0$, where the homogeneous solution $\hat{u} := u - v$ is obtained by subtracting the particular solution $v$ from the general solution $u$ as discussed in [33]. However, in our settings we do not know either the locations nor the magnitudes of the point sources and therefore, we cannot apply this technique.

- Similar to other data-driven approaches, the PIBI-Net solution relies strongly on the quality of the observations. If the observations do not cover the important regions for the dynamics, the problem might be ill-posed. Therefore, data-driven approaches cannot compete with solutions that contain more information through boundary conditions or known sources. As a result, neither the PIBI-Net nor the PINN approach can adequately cover areas such as the lower right corner on their own, due to the lack of observations in the experiment depicted in Fig. 2.

- Since PIBI-Net only requires first-order derivatives, it enables us to use a broader range of MLP models.

- As the integral calculation of PIBI-Net is based on the Monte Carlo integration, a reasonable amount of integral points is therefore required to obtain valid results. On the one hand, too few integration points would lead to a wrong calculation of the integral. On the other hand, too many integration points can be at the expense of computing efficiency.

- In our real-world experiment, we fit groundwater surfaces as illustrated in Fig. 4. We showcased the effectiveness of reformulating point sources through Eq. (12) within physics-informed deep learning approaches. However, in scenarios involving non-homogeneous PDEs with non-point sources, we would lose PIBI-Net's advantage of being able to reduce the dimension by one. This is because the integral of the source potential given by Eq. (6) must be computed across the entire domain, which becomes notably time and computationally intensive when the right-hand side term of the Poisson equation remains unknown.

## 5. Conclusion

In summary, our proposed PIBI-Net method is a data-driven deep learning approach that inherently satisfies the solution of established PDEs like the Laplace or Poisson equation without requiring boundary conditions, while assimilating data. In the two-dimensional toy examples we have shown that PIBI-Net clearly outperforms PINNs in terms of accuracy, while requiring only integration points at the boundary. This dimension reduction property is highly advantageous in many problem settings, as most data-driven real-world applications are in three or four dimensions. Moreover, training PIBI-Net in data assimilation settings without a specific focus on the boundary is more efficient compared to PINN, as it does not require any loss weighting. In the real-world steady-state groundwater application, we additionally demonstrated that PIBI-Net effectively solve inverse problems with point sources. Consequently, PIBI-Net can be used for all dynamical systems that are based on a linear PDE with constant coefficients. For unknown right-hand side terms of the Poisson equation that are not point sources, however, an integral over the entire domain must still be solved. Therefore, further research needs to be done in this direction.

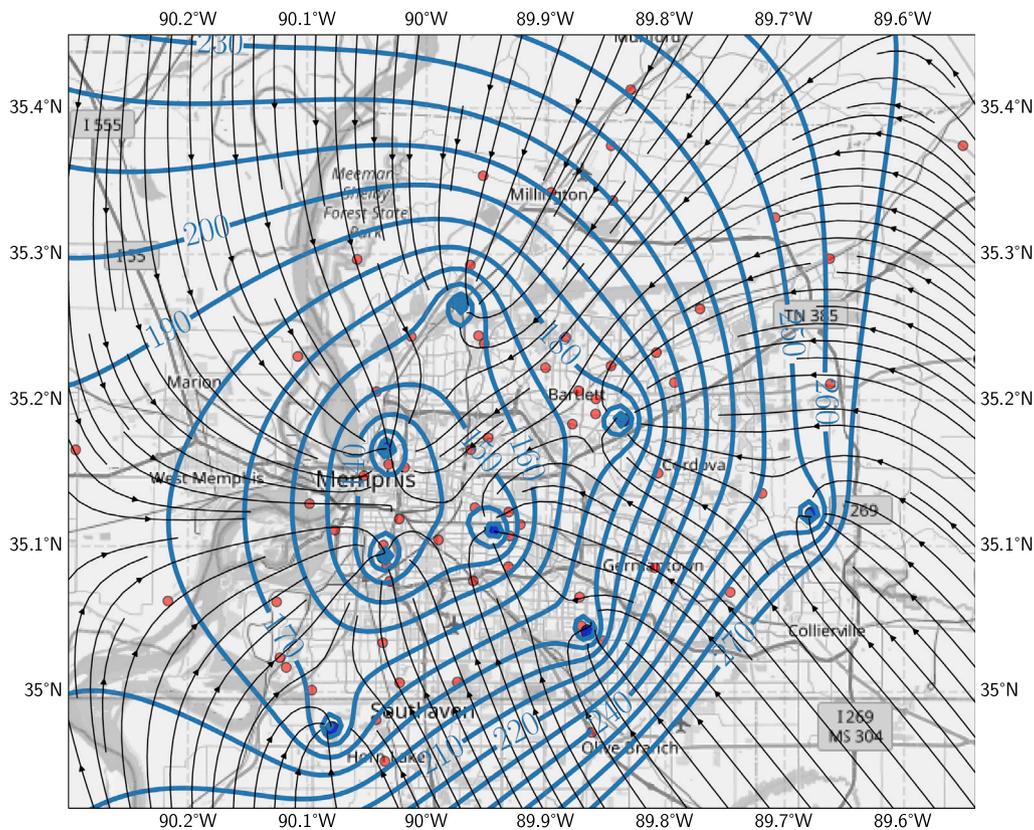

**Fig. 4.** Visualisation of the water head contour lines based on the well measurements indicated as red dots. The black lines indicate the streamlines of the groundwater flows and the blue dots show the fitted point sources, which represent the pumping wells.

**CRediT authorship contribution statement**

**Monika Nagy-Huber:** Conceptualisation, Data curation, Investigation, Formal analysis, Methodology, Validation, Visualisation, Writing – original draft, Writing – review & editing. **Volker Roth:** Supervision, Project administration, Writing – review & editing, Data curation, Resources.

**Declaration of competing interest**

The authors declare that they have no known competing financial interests or personal relationships that could have appeared to influence the work reported in this paper.

**Data and code availability**

Data from Kingsbury [1] was used for the groundwater simulation. The implementations that generated the presented results are available on our Github repository: https://github.com/MonikaNagy-Huber/PIBI-Net.git.

**Acknowledgements**

We thank Remo von Rickenbach, Fabricio Arend Torres, Jonathan Aellen and Marcello Negri for helpful discussions.

**References**

[1] J.A. Kingsbury, Altitude of the Potentiometric Surface, 2000–15, and Historical Water-Level Changes in the Memphis Aquifer in the Memphis Area, Tennessee, Technical Report, US Geological Survey, 2018, http://dx.doi.org/10.3133/sim3415.

[2] A. Khan, D.A. Lowther, Physics informed neural networks for electromagnetic analysis, IEEE Trans. Magn. 58 (9) (2022) 1–4, http://dx.doi.org/10.1109/TMAG.2022.3161814.

[3] J.D. Anderson, G. Degrez, E. Dick, R. Grundmann, Computational Fluid Dynamics: an Introduction, Springer Science & Business Media, 2013, http://dx.doi.org/10.1007/978-3-540-85056-4.

[4] X. Xun, J. Cao, B. Mallick, A. Maity, R.J. Carroll, Parameter estimation of partial differential equation models, J. Amer. Statist. Assoc. 108 (503) (2013) 1009–1020, http://dx.doi.org/10.1080/01621459.2013.794730.

[5] R. Arcucci, J. Zhu, S. Hu, Y.-K. Guo, Deep data assimilation: integrating deep learning with data assimilation, Appl. Sci. 11 (3) (2021) 1114, http://dx.doi.org/10.3390/app11031114.

[6] C. Buizza, C.Q. Casas, P. Nadler, J. Mack, S. Marrone, Z. Titus, C. Le Cornec, E. Heylen, T. Dur, L.B. Ruiz, et al., Data learning: integrating data assimilation and machine learning, J. Comput. Sci. 58 (2022) 101525, http://dx.doi.org/10.1016/j.jocs.2021.101525.

[7] S. Cheng, C. Quilodrán-Casas, S. Ouala, A. Farchi, C. Liu, P. Tandeo, R. Fablet, D. Lucor, B. Iooss, J. Brajard, et al., Machine learning with data assimilation and uncertainty quantification for dynamical systems: a review, IEEE/CAA J. Autom. Sin. 10 (6) (2023) 1361–1387, http://dx.doi.org/10.1109/JAS.2023.123537.

[8] R. Nussbaumer, S. Bauer, L. Benoit, G. Mariethoz, F. Liechti, B. Schmid, Quantifying year-round nocturnal bird migration with a fluid dynamics model, J. R. Soc. Interface 18 (179) (2021) 20210194, http://dx.doi.org/10.1098/rsif.2021.0194.

[9] M. Asch, M. Bocquet, M. Nodet, Data Assimilation: Methods, Algorithms, and Applications, SIAM, 2016, http://dx.doi.org/10.1137/1.9781611974546.

[10] C. Liu, R. Fu, D. Xiao, R. Stefanescu, P. Sharma, C. Zhu, S. Sun, C. Wang, EnKF data-driven reduced order assimilation system, Eng. Anal. Bound. Elem. 139 (2022) 46–55, http://dx.doi.org/10.1016/j.enganabound.2022.02.016.

[11] M. Raissi, P. Perdikaris, G.E. Karniadakis, Physics-informed neural networks: A deep learning framework for solving forward and inverse problems involving nonlinear partial differential equations, J. Comput. Phys. 378 (2019) 686–707, http://dx.doi.org/10.1016/j.jcp.2018.10.045.

[12] L. Yang, X. Meng, G.E. Karniadakis, B-PINNs: Bayesian physics-informed neural networks for forward and inverse PDE problems with noisy data, J. Comput. Phys. (ISSN: 0021-9991) 425 (2021) 109913, http://dx.doi.org/10.1016/j.jcp.2020.109913.

[13] S. Cuomo, V.S. Di Cola, F. Giampaolo, G. Rozza, M. Raissi, F. Piccialli, Scientific machine learning through physics-informed neural networks: where we are and what's next, J. Sci. Comput. 92 (3) (2022) 88, http://dx.doi.org/10.1007/s10915-022-01939-z.

[14] P. Sharma, W.T. Chung, B. Akoush, M. Ihme, A review of physics-informed machine learning in fluid mechanics, Energies 16 (5) (2023) 2343, http://dx.doi.org/10.3390/en16052343.

[15] K. Tang, X. Wan, C. Yang, DAS-PINNs: A deep adaptive sampling method for solving high-dimensional partial differential equations, J. Comput. Phys. 476 (2023) 111868, http://dx.doi.org/10.1016/j.jcp.2022.111868.

[16] J. Fu, D. Xiao, R. Fu, C. Li, C. Zhu, R. Arcucci, I.M. Navon, Physics-data combined machine learning for parametric reduced-order modelling of nonlinear dynamical systems in small-data regimes, Comput. Methods Appl. Mech. Engrg. 404 (2023) 115771, http://dx.doi.org/10.1016/j.cma.2022.115771.

[17] L. Cheng, E.A. Illarramendi, G. Bogopolsky, M. Bauerheim, B. Cuenot, Using neural networks to solve the 2D Poisson equation for electric field computation in plasma fluid simulations, 2021, arXiv preprint arXiv:2109.13076. https://arxiv.org/abs/2109.13076.

[18] A.G. Özbay, A. Hamzehloo, S. Laizet, P. Tzirakis, G. Rizos, B. Schuller, Poisson CNN: Convolutional neural networks for the solution of the Poisson equation on a cartesian mesh, Data-Centric Eng. 2 (2021) e6, http://dx.doi.org/10.1017/dce.2021.7.

[19] T. Shan, W. Tang, X. Dang, M. Li, F. Yang, S. Xu, J. Wu, Study on a fast solver for Poisson's equation based on deep learning technique, IEEE Trans. Antennas and Propagation 68 (9) (2020) 6725–6733, http://dx.doi.org/10.1109/TAP.2020.2985172.

[20] J. Tompson, K. Schlachter, P. Sprechmann, K. Perlin, Accelerating eulerian fluid simulation with convolutional networks, in: International Conference on Machine Learning, PMLR, 2017, pp. 3424–3433, https://proceedings.mlr.press/v70/tompson17a.html.

[21] S. Markidis, The old and the new: Can physics-informed deep-learning replace traditional linear solvers? Front. Big Data 4 (2021) 669097, http://dx.doi.org/10.3389/fdata.2021.669097.

[22] A.N. Maria Antony, N. Narisetti, E. Gladilin, FDM data driven U-Net as a 2D Laplace PINN solver, Sci. Rep. 13 (1) (2023) 9116, http://dx.doi.org/10.1038/s41598-023-35531-8.

[23] R.I. Abdulkadirov, P.A. Lyakhov, N.N. Nagornov, Solving Poisson equation by physics-informed neural network with natural gradient descent with momentum, in: 2023 Seminar on Signal Processing, IEEE, 2023, pp. 3–6, http://dx.doi.org/10.1109/IEEECONF60473.2023.10366065.

[24] P.-H. Chiu, J.C. Wong, C. Ooi, M.H. Dao, Y.-S. Ong, CAN-PINN: A fast physics-informed neural network based on coupled-automatic–numerical differentiation method, Comput. Methods Appl. Mech. Engrg. 395 (2022) 114909, http://dx.doi.org/10.1016/j.cma.2022.114909.

[25] O. Steinbach, Numerical Approximation Methods for Elliptic Boundary Value Problems: Finite and Boundary Elements, Springer Science & Business Media, 2007, http://dx.doi.org/10.1007/978-0-387-68805-3.

[26] J. Sun, Y. Liu, Y. Wang, Z. Yao, X. Zheng, BINN: A deep learning approach for computational mechanics problems based on boundary integral equations, Comput. Methods Appl. Mech. Engrg. 410 (2023) 116012, http://dx.doi.org/10.1016/j.cma.2023.116012.

[27] G. Lin, F. Chen, P. Hu, X. Chen, J. Chen, J. Wang, Z. Shi, BI-GreenNet: learning green's functions by boundary integral network, Commun. Math. Stat. 11 (1) (2023) 103–129, http://dx.doi.org/10.1007/s40304-023-00338-6.

[28] R. Villalpando-Vizcaino, B. Waldron, D. Larsen, S. Schoefernacker, Development of a numerical multi-layered groundwater model to simulate inter-aquifer water exchange in shelby county, Tennessee, Water 13 (18) (2021) 2583, http://dx.doi.org/10.3390/w13182583.

[29] B. Malgrange, Existence et approximation des solutions des équations aux dérivées partielles et des équations de convolution, in: Annales de l'institut Fourier, Vol. 6, 1956, pp. 271–355, http://dx.doi.org/10.5802/aif.65.

[30] L. Ehrenpreis, Solution of some problems of division: Part i. division by a polynomial of derivation, Amer. J. Math. 76 (4) (1954) 883–903, http://dx.doi.org/10.2307/2372662.

[31] S. Kesavan, A. Vasudevamurthy, On some boundary element methods for the heat equation, Numer. Math. 46 (1985) 101–120, http://dx.doi.org/10.1007/BF01400258.

[32] D.M. Misljenovic, Boundary element method and wave equation, Appl. Math. Model. 6 (3) (1982) 205–208, http://dx.doi.org/10.1016/0307-904X(82)90012-9.

[33] M.A. Golberg, The method of fundamental solutions for Poisson's equation, Eng. Anal. Bound. Elem. 16 (3) (1995) 205–213, http://dx.doi.org/10.1016/0955-7997(95)00062-3.

[34] A. Paszke, S. Gross, F. Massa, A. Lerer, J. Bradbury, G. Chanan, T. Killeen, Z. Lin, N. Gimelshein, L. Antiga, et al., Pytorch: An imperative style, high-performance deep learning library, Adv. Neural Inf. Process. Syst. 32 (2019) https://proceedings.neurips.cc/paper_files/paper/2019/file/bdbca288fee7f92f2bfa9f7012727740-Paper.pdf.

[35] M. Abadi, P. Barham, J. Chen, Z. Chen, A. Davis, J. Dean, M. Devin, S. Ghemawat, G. Irving, M. Isard, et al., Tensorflow: a system for large-scale machine learning, in: Osdi, Vol. 16, Savannah, GA, USA, 2016, pp. 265–283, https://dl.acm.org/doi/10.5555/3026877.3026899.

[36] D.P. Kingma, J. Ba, Adam: A method for stochastic optimization, 2014, arXiv preprint arXiv:1412.6980. https://arxiv.org/abs/1412.6980.

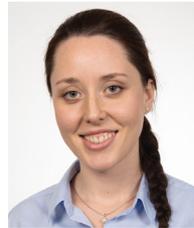

**Monika Nagy-Huber** is a Ph.D. student in Computer Science at the University of Basel. She holds an M.Sc. in Mathematics with focus on numerics as well as algebra and number theory. Her research is focusing on data-driven physics-informed machine learning for solving partial differential equations in dynamical systems.

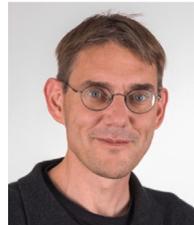

**Volker Roth** has studied Physics, and he received his Ph.D. degree in Computer Science from the University of Bonn in 2001. After some years as a postdoc at ETH Zurich, he joined the University of Basel in 2007. His main research interests include machine learning, statistical models for data analysis and biomedical applications.